# Hybrid Model for Solving Multi-Objective Problems Using Evolutionary Algorithm and Tabu Search

| Rjab Hajlaoui | Mariem Gzara | Abdelaziz Dammak |
|---|---|---|
| Department of computer Sciences | Department of computer Sciences | Department of Applied Quantitative Methods |
| Hail University | Monastir University | Sfax University |
| Hail, KSA | Monastir, Tunisia | Sfax, Tunisia |
| Hajlaoui.rjab@gmail.com | Mariem.gzara@isimsf.rnu.tn | abdelaziz.dammak@fsegs.rnu.tn |

*Abstract*—This paper presents a new multi-objective hybrid model that makes cooperation between the strength of research of neighborhood methods presented by the tabu search (TS) and the important exploration capacity of evolutionary algorithm. This model was implemented and tested in benchmark functions (ZDT1, ZDT2, and ZDT3), using a network of computers.

## I. INTRODUCTION

The resolution of optimization problems requires satisfaction of several contradictory objectives. The optimization of one objective can be at the expense of others. In this case, the important number of constraints, the behaviour of these objectives' functions, as well as the landscape of the Pareto Front (Convexity, discontinuities, and multi modes) put some inherent difficulties to overcome [1].

Mainly, metaheuristics are considered as high-level algorithmic strategies that are used to guide other heuristics or algorithms in their search for the space to feasible solutions of the optimal value (in the single-objective case) and the set of optimal values (in the multi-objective case).

Using parallel metaheuristics (tabu search method, genetic algorithms, simulated annealing, etc.) permits toexploit the specific advantages of every approach and to establish a new record to find better solutions [2].

In this optic, we propose a new cooperative model that directs search in the less explored zones and that visits the entire feasible surface. This parallel cooperative model is based on three basic components that evolve and cooperate simultaneously. These components are: Strength Pareto Evolutionary Algorithm (SPEA), Intensificator Tabu Search (ITS), and Diversificator Tabu Search (DTS).

This paper is divided into four sections. Sections 1 and 2 simultaneously describe the Evolutionary Algorithm andthe Tabu Search. Section 3 presents our cooperative model between the elitist Evolutionary Algorithm (SPEA) and the Tabu Search for multi-objective optimization.

Finally, section 4 gives experimental results and discusses the performance of our model.

## II. STRENGTH PARETO EVOLUTIONARY ALGORITHM

Evolutionary algorithms seem particularly suitable to solve multi-objective optimization problems because they deal simultaneously with a set of possible solutions (the so-called population). This allows to find several members of the Pareto optimal set in a single "run" of the algorithm, instead of having to perform a series of separate runs as in the case of the traditional mathematical programming techniques. In addition, evolutionary algorithms are less susceptible to the shape or continuity of the Pareto front; e.g, they can easily deal with discontinuous or concave Pareto fronts.

The choice of the SPEA method is not arbitrary. Indeed, many comparative studies [3] have been achieved on different multi-objective evolutionary algorithms and have shown that the SPEA method is clearly superior. This last method integrates only one algorithm technique established by the EA with new techniques in order to find optimal parallel solutions.

The SPEA is simultaneously based on concepts of no dominance and of elitism. In addition to an initial population of size N, the SPEA uses an external population—called archive—of size M to maintain the pareto optimal solutions. In this method, the passage from one generation to another starts with the updating of the archive. All non-dominated individuals are copied in the archive and the dominated individuals already present are deleted. If the number of individuals in the archive exceeds a given number, a clustering technique is performed to reduce the archive. The performance of every individual is then updated before doing the selection while using the two groups. To finish, one applies the genetic operators of crossing and mutation [3].

ALGORITHM 1: SPEA ALGORITHM

Step 1: Generate random initial population P and create the empty external set of non-dominated individuals P0.

Step 2: Evaluate objective functions for each individual in P.

Step 3: Copy non-dominated members of P to Pa.

Step 4: Remove solutions within Pa which are covered by any other member of Pa.

Step 5: If the number of externally kept non-dominated solutions exceeds a given size, then use the clustering technique.

Step 6: Calculate the fitness of each individual in P, as well as, in Pa.

Step 7: Select individuals from P U Pa, until the mating pool is filled.

Step 8: Apply recombination and mutation to members of the mating pool in order to create a new population P.

Step 9: If a maximum number of generations is reached, then stop, else go to Step 2.

## III. TABU SEARCH

Tabu search (TS) is a metaheuristic procedure introduced by Glover [4]. It is used for a local search heuristic algorithm to explore the space of solutions beyond the simple local optimum. It has been applied to a wide range of practical optimization applications, producing an accelerated growth of Tabu search algorithms in recent years.

A tabu list is a set of solutions determined by historical information from the last iterations of the algorithm. At each iteration, given the current solution x and its corresponding neighborhood N(x), the procedure moves to the solution in the neighborhood N(x) that mostly improves the objective function. However, moves that lead to solutions on the tabu list are forbidden or are tabu. If there are no improving moves, TS chooses the move which at least changes the objective function value. The tabu list avoids returning to the local optimum from which the procedure has recently escaped. A basic element of the tabu search is the aspiration criterion, which determines when a move is admissible despite being on the tabu list. One termination criterion for the tabu procedure is a limit in the number of consecutive moves for which no improvement occurs.

ALGORITHM 2: TABU SEARCH ALGORITHM

Step 1: Choose an initial solution in S (search space). Set i * = i and k = 0

Step 2: Apply k = k+1 and generate a subset of solutions in N(i,k) so that:

The tabu movements are not chosen

The aspiration criterion a(i,m ) is applied

N(i,k) is the neighborhood of the current solution i at iteration k.

Step 3: Choose the best solution i' among N(i,k), then apply i = better i'

Step 4: If f(i) <= f(i *), then apply i * = i

Step 5: Update the list T and aspiration criterion.

Step 6: If a stop condition is reached, then stop.

Otherwise, return to Step 2.

## IV. COOPERATIVE MODEL: EVOLUTIONARY ALGORITHM-TABU SEARCH FOR THE MULTI-OBJECTIVE OPTIMIZATION(COMOEATS):

Our goal is to exploit the advantages of EA and TS to achieve a level-headed research that leads to a uniform distribution which covers the totality of the pareto optimal solutions. This search avoids the premature convergence and the omission of promising solutions at the same time.

Fig.1. shows two possible distributions where the solutions given are limited on a short part of the front and the choice of decider is reduced. To overcome this problem, we are developing a new hybrid model in order to find an unexplored zone to which the search is directed (using Diversificator tabu search). After that an intensificator tabu search is applied in Fig. 3.

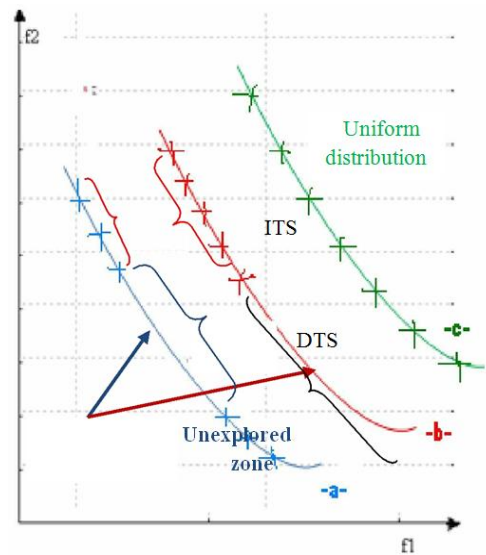

Fig.1. Example of distributions

The working of the model is illustrated as follows:

ALGORITHM 3 : COMOEATS ALGORITHM

Step 1: Choose an initial population

Step 2: Execute the SPEA method on time

Step 3: In parallel with SPEA, casting an intensificator tabu search with a solution of the front and another diversificator with all solutions of the front

Step 4: The improved solutions obtained will be integrated in the population archives (population of departure for the next iteration of the SPEA)

Step 5: If a stop condition is reached (number of generations), stop. Else, return to step 3.

The model that we propose works in three parallel components:

- The first uses a multi-objective evolutionary algorithm (SPEA) that, each time, generates a Pareto front increasingly improved. This algorithm performs global search and discovers multiple optimal solutions.

- An ITS, receives a non-dominated solution taken at random from the Pareto Front generated by the SPEA algorithm to be the topic of improvement. For each iteration, the neighborhood of the current solution is generated by keeping only non-dominated solutions that belong to the intensification zone fig 2. There are two cases:

*The neighborhood includes a solution dominating S0, which, in this case, will be the new solution of departure.

*There isn't any solution that dominates S0, and in this case a solution from neighborhood will be taken as the one of departure. These stages will be reiterated until a maximum number of iterations are reached.

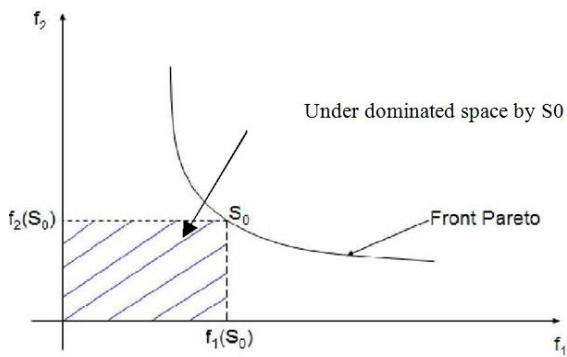

Fig.2: Space of research for the intensificator TS

- The DTS detects a less explored zone of the search space and performs a local search in order to discover new solutions.

In the bi-objective case, the DTS detects the two most distant and successive points of the pareto front (either SL1 or SL2), calculates the middle vector cost of these points (Cm), and then marks the best solution belonging to the hatched dominant zone Cm , fig.3.

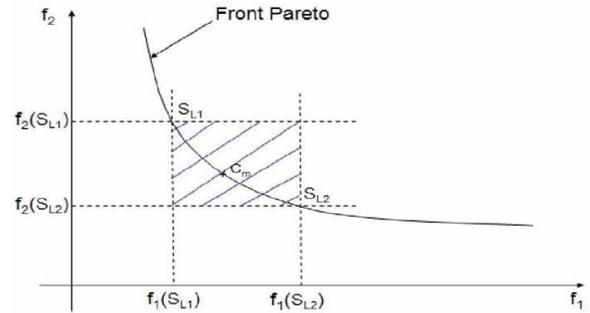

Fig.3: Space of research for the diversificateur TS

For all individuals generated by the DTS and ITS, a mechanism of selection decides that a new improved individual must be introduced in the population archives. This last will be taken as an initial solution for the SPEA method in the following iteration.

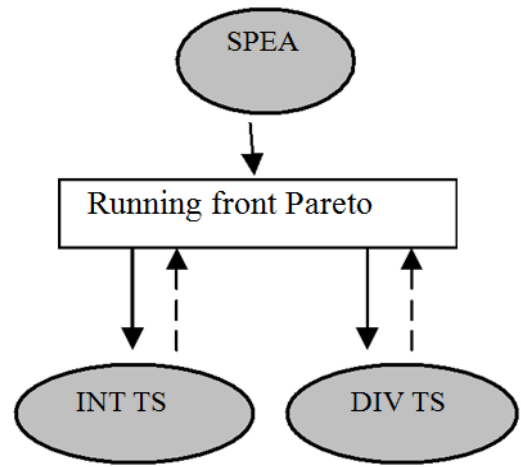

Fig.4: Diagram of basis of the COMOEATS model

V.    EXPERIMENTAL RESULTS:

Our COMOEATS model has been programmed with Visual C++, executed on computer network using the MPI library (Message Passing Interface).

For each 10 iterations (determined by user) of SPEA algorithm, the solutions of Pareto front obtained will be improved by the Intensificator and Diversificatu tabu search. Then, new solutions will be integrated in the next population.

The parameters of the method are illustrated in the following table:

| SPEA | TS |
|---|---|
| -size of the initial population: 100 individuals | -size tabu list: 50 |
| -size of the population archives: 100 individuals | -number of iterations: 200 |
| - number of generations: 400 | -Tabu life: 50 |
| - probability of crossover: 0.9 | |
| -probability of mutation: 0.01 | |
| Binary -coding: 16 bits | |

We compare our COMOEATS method to SPEA method (experimentations are done on bi-objective test functions: ZDT1, ZDT2, and ZDT3). Then, we present and discuss the Pareto Front generated by every method for each function [7].

The following table presents a set of experimentations achieved on benchmarks. We use the following metrics:

Contribution: It evaluates the proportion of Pareto solution brought by each of the two foreheads F1 and F2 [5].

Entropy: The entropy uses the notion of nest to evaluate the distribution of solutions on the Pareto front. Closer to 1 the gotten value is, better is the distribution.

Spacing: This metric is based on a calculation of the distance between solutions. It permits to measure the uniformity of the distribution of points of the compromise surface in the space of objectives [6]. Used with other measures, it gives an interesting indication on the convergence of the method used.

Metric S: This metric uses a reference point while calculating the hyper-volume of the multi-dimensional region comprised between the Pareto front and the reference point [6].

-ZDT1: this problem has a convex POF

TABLE.2: METRICS FOR ZDT1

| | COMOEATS | SPEA |
|---|---|---|
| Contribution | 0 .507042 | 0.492958 |
| Entropie | 0.367399 | 0.360803 |
| Spacing | 0.0256606 | 0.0203861 |
| Métrique S | 0.55787 | 0.5524335 |

According to this representation, we notice an advantage of our model in relation to SPEA at the level of the two metrics contribution and entropy. This difference implies a better distribution of solutions on the Pareto front.

This improvement became clearer in fig.5, where we succeeded to visit unexplored zones by SPEA and get various solutions. On the other hand, our hybrid model bust a great number of solutions that cover all the front, thence the hyper-volume hand from 0.5524335 to 0.55787.

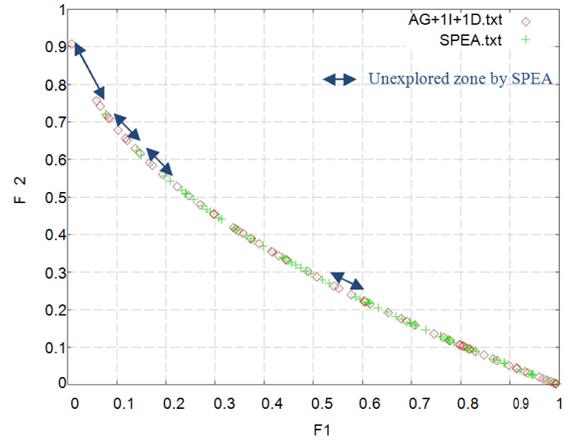

Fig. 5: Distribution of solutions for the two models COMOEATS and SPEA (Case of ZDT1).

-ZDT2: this problem has a non-convex POF:

TABLE.3: METRICS FOR ZDT2

| | COMOEATS | SPEA |
|---|---|---|
| Contribution | 0.507092 | 0.492958 |
| Entropie | 0.371775 | 0.360803 |
| Spacing | 0.0276606 | 0.0203861 |
| Métrique S | 0.55689 | 0.5524335 |

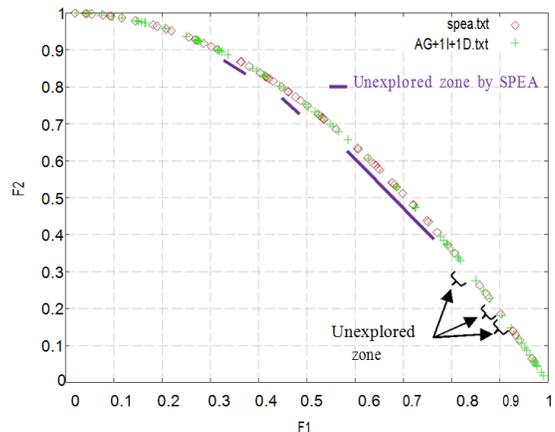

Fig. 6: Distribution of solutions for the two models COMOEATS and SPEA (Case of ZDT2)

For this type of problems, we notice a remarkable improvement particularly at the level of new solutions discovered by our model. At the same time, there remain some less discovered parts. For this reason, we must use several Diversificator tabu searches.

-ZDT3: this problem has a discontinuous POF:

<p align="center">Table.4 : Metrics for ZDT3</p>

|  | COMOEATS | SPEA |
|---|---|---|
| Contribution | 0.507042 | 0.496454 |
| Entropie | 0.373243 | 0.365199 |
| Spacing | 0.0116797 | 0.0206785 |
| Métrique S | 0.741164 | 0.750998 |

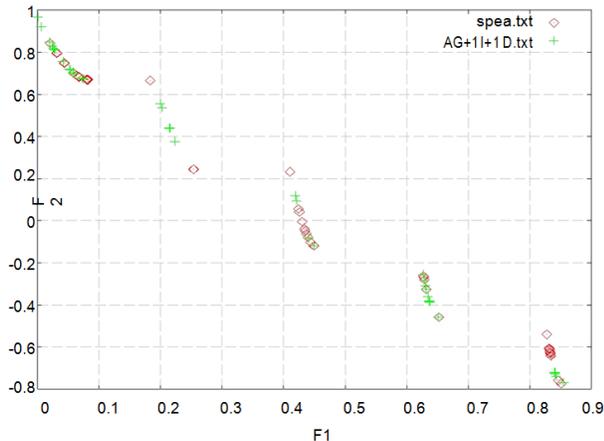

Fig. 7: Distribution of solutions for the two models COMOEATS and SPEA (Case of ZDT3)

For problems having a discontinuous forehead, our method proves to be interesting at the level of all metrics. This improvement became lucid in Fig.7 which presents a good uniformity.

## VI. Conclusion

In this paper, we presented a hybrid approach (COMOEATS) for the resolution of multi-objective problems. This approach is based on elitist evolutionary algorithm (SPEA), Intensificator Tabu search (ITS), and Diversificator tabu search (DTS). A set of experimentations have been conducted in benchmark functions known in the literature. The achieved results show a clear improvement of our approach compared to that using SPEA. The future tasks will be carried on the use of several tabu searches, hence the generalization of the method with multi-objective problems.